\documentclass{article}

\usepackage[preprint]{neurips_2026}

\usepackage[utf8]{inputenc} 
\usepackage[T1]{fontenc}    
\usepackage{hyperref}       
\usepackage{url}            
\usepackage{booktabs}       
\usepackage{amsfonts}       
\usepackage{nicefrac}       
\usepackage{microtype}      
\usepackage{xcolor}         

\usepackage{amsmath}
\usepackage{graphicx}
\usepackage{subcaption}
\usepackage{multirow}
\usepackage[normalem]{ulem}
\usepackage{soul}

\usepackage{algorithm}
\usepackage{algpseudocode}
\usepackage{wrapfig}
\usepackage{enumitem}
\usepackage[table]{xcolor} 

\newtheorem{Theorem}{Theorem}
\newtheorem{Proposition}[Theorem]{Proposition}

\newcommand{\boldhdr}[1]{\vspace{0.05cm}\noindent\textbf{#1}.}

\newcommand{\rateinline}[2]{#1\color{gray}{\tiny$\pm$ #2}}

\title{Consistently Informative Soft-Label Temperature for Knowledge Distillation}

\author{%
Hoang-Chau Luong$^{1,\dagger}$ \quad Nghia Van Vo$^{2,\dagger}$ \quad Kaiqi Zhao$^2$ \quad Lingwei Chen$^1$\\
$^1$Rochester Institute of Technology, Rochester, NY, USA \\
$^2$Oakland University, Rochester, MI, USA\\
$^\dagger$These authors contributed equally\\
\texttt{cl6300@rit.edu, \{nghiavo, kaiqizhao\}@oakland.edu, lwcics@rit.edu}
}
\begin{document}

\maketitle

\begin{abstract}
    Knowledge distillation (KD) transfers knowledge from a high-capacity teacher to a compact student by matching their predictive distributions, with temperature scaling $\tau$ serving as a central mechanism for smoothing teacher predictions and exposing informative ``dark knowledge'' beyond the hard label. However, the standard fixed-temperature design is inherently sample-agnostic. Since samples differ in logit scale and learning difficulty, a single global temperature produces teacher soft labels with highly inconsistent entropy: some predictions remain overly sharp and provide limited inter-class information, whereas others become over-smoothed and lose class-discriminative information. Moreover, sharing the same temperature between teacher and student further imposes rigid logit-scale alignment despite their capacity mismatch.
    To address these limitations, we propose \textbf{CIST} (\textbf{C}onsistently \textbf{I}nformative \textbf{S}oft-label \textbf{T}emperature), which assigns separate sample-wise adaptive temperatures to the teacher and student. This design produces consistently informative teacher soft labels while relaxing rigid teacher--student logit-scale matching. It also reweights the distillation objective according to teacher confidence and student learning difficulty. Theoretically, we show that teacher-label entropy is largely governed by the ratio between the maximum teacher logit and the temperature, providing a principled basis for adaptive smoothing. Empirically, CIST mitigates the inconsistency induced by fixed temperature, and experiments on both vision and language distillation tasks show consistent improvements over standard KD and strong baselines with negligible computational overhead.
\end{abstract}

\vspace{-0.2cm}
\section{Introduction}
\vspace{-0.1cm}

Deep neural networks (DNNs) have achieved state-of-the-art performance across a wide range of tasks, largely driven by the continual scaling of model capacity. However, this improvement often comes with substantial computational and memory costs, making large models difficult to deploy in resource-constrained environments such as mobile and edge devices. To address this challenge, model compression has been extensively studied. Among existing compression techniques, knowledge distillation (KD)~\citep{hinton2015distilling} has emerged as a simple and effective approach, where a compact student model learns from a high-capacity teacher by mimicking its predictive behavior~\citep{romero15-iclr,cho2019efficacy,gou2021knowledge}.

\begin{figure}[t]
\centering
    \centering
    \includegraphics[width=0.95\linewidth]{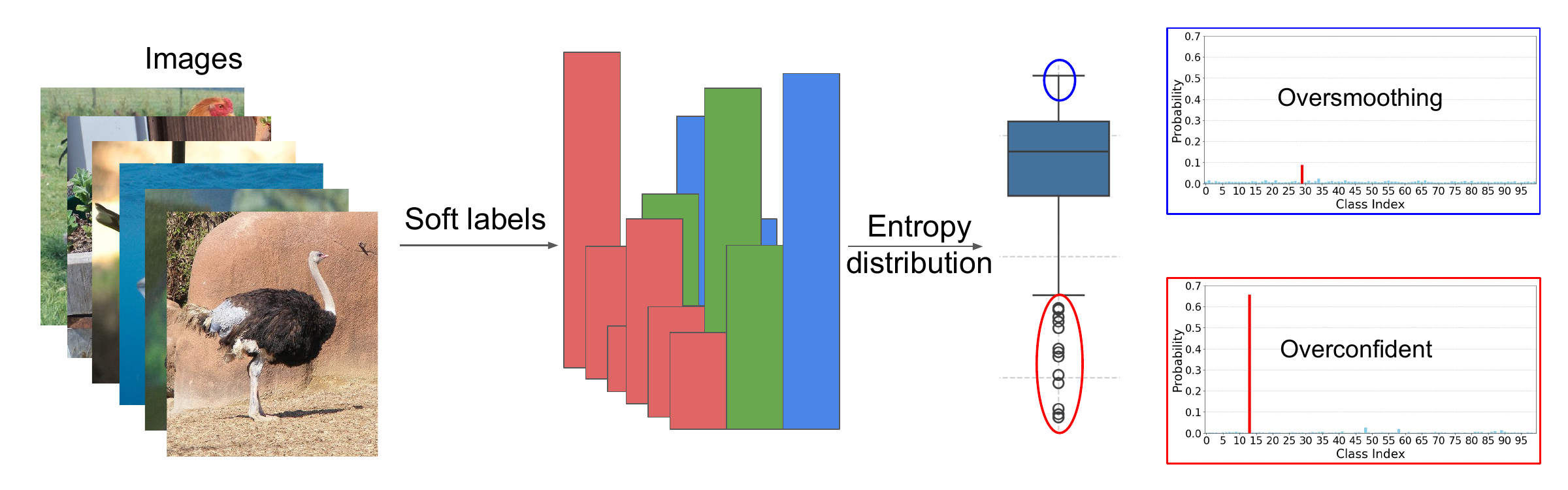}
    \caption{Teacher soft label generation in KD. With fixed temperature, some samples remain overconfident and uninformative (\textcolor{red}{low entropy}), while others are oversmoothed (\textcolor{blue}{high entropy}). These inappropriately smoothed samples can reduce knowledge transfer.}
    \vspace{-0.5cm}
    \label{fig:problem}
\end{figure}

A central mechanism in KD is temperature $\tau$, which divides teacher and student logits by a temperature hyperparameter before the softmax operation. By increasing $\tau$, the teacher distribution becomes less peaked and exposes inter-class relationships, commonly known as ``dark knowledge''~\citep{hinton2015distilling,tang2020understanding,zhao2022decoupled,wei2024dynamic}. In standard KD, however, $\tau$ is selected as a single global constant. This choice implicitly assumes that all teacher predictions require the same amount of smoothing, even though their logit scales can vary substantially across samples.
We argue that this assumption is problematic because the entropy of the softened teacher distribution is highly sensitive to the ratio between the dominant logit and the temperature. 
When the dominant logit is large relative to $\tau$, the softened label remains sharply concentrated and provides little non-target information. 
Conversely, when the logit gaps are small relative to $\tau$, the distribution can become excessively flat, weakening class-discriminative signals. Figure~\ref{fig:problem} illustrates this issue: under the same fixed temperature, teacher soft-label entropy varies widely across samples.

This entropy inconsistency reveals that samples do not contribute uniformly to distillation: some provide informative soft supervision, while others provide unreliable training signals. Standard KD assigns the same loss weight to all samples, although teacher predictions differ in confidence and the student may learn from different samples at different rates. Low-confidence teacher predictions can provide unreliable soft targets, while hard samples for the current student may lead to noisy or unstable gradients. This motivates a curriculum-based distillation strategy, where training emphasizes samples that provide reliable and informative supervision~\citep{bengio2009curriculum,li2023curriculum}. Although prior work has explored curriculum temperatures~\citep{li2023curriculum}, logit standardization before softmax~\citep{sun2024logit}, and entropy-based loss weighting~\citep{su2025ea}, they do not directly address the entropy inconsistency of teacher soft labels under fixed temperature. In addition, standard KD commonly shares the same temperature between teacher and student, which can implicitly encourage rigid logit-scale matching despite their capacity mismatch~\citep{sun2024logit}.

To address these limitations, we propose \textbf{CIST} (\textbf{C}onsistently \textbf{I}nformative \textbf{S}oft-label \textbf{T}emperature), an effective distillation framework that combines consistent informative soft labels with confidence-aware curriculum regularization. It first assigns sample-wise adaptive temperatures to the teacher based on the dominant teacher logits, stabilizing teacher-label entropy and producing consistently informative soft supervision. It then applies separate adaptive temperatures to the teacher and student, relaxing the rigid logit-scale matching imposed by shared-temperature KD. Finally, CIST reweights the distillation loss according to teacher confidence and student learning difficulty, emphasizing reliable and learnable soft targets while down-weighting uncertain or poorly learned samples.
Theoretically, we show that teacher-label entropy is largely governed by the ratio between the maximum teacher logit and the temperature, providing a principled basis for our adaptive design.
We validate CIST on both vision and language distillation tasks. On CIFAR-100 and ImageNet, CIST consistently improves standard KD across diverse teacher--student architectures. On instruction-following language distillation, CIST also outperforms other distillation baselines across multiple teacher--student pairs and evaluation benchmarks. Overall, CIST achieves strong performance compared with competitive distillation baselines while introducing negligible computational overhead.

In summary, our contributions are threefold:
\begin{itemize}[leftmargin=20pt]
    \vspace{-0.1cm}
    \item We show a key limitation of fixed-temperature KD: a single global temperature induces inconsistent teacher-label entropy across samples, leading to uninformative soft labels.

    \item We propose \textbf{CIST}, a logit-based KD framework motivated by an entropy-based analysis of teacher soft labels. By combining sample-wise temperature, independent teacher--student temperature, and confidence-aware curriculum regularization, CIST produces informative soft targets, relaxes rigid logit-scale alignment, and emphasizes reliable distillation signals.

    \item We validate CIST across vision and language distillation tasks, where it consistently outperforms strong baselines on CIFAR-100, ImageNet, and instruction-following language distillation.
\end{itemize}

\section{Related Works}
\label{sec:related_works}

\textbf{Knowledge distillation} aims to transfer the dark knowledge from a high-capacity teacher model to a lightweight student model. By learning from the soft labels produced by the teacher, the student often achieves better generalization than when trained solely on hard labels. Traditionally, KD trains the student by minimizing the Kullback–Leibler (KL) divergence between the teacher's and student's predicted probability distributions. These probabilities are obtained by applying the softmax function to the models' respective logits. KD methods can broadly be categorized into three groups: logit-based approaches~\citep{hinton2015distilling,zhao2022decoupled,jin2023multi,sun2024logit,zheng2024knowledge}, which directly match the output distributions; feature-based methods~\citep{romero15-iclr,zagoruyko2017paying,park2019relational,tian2019contrastive,heo2019comprehensive,chen2021distilling}, which align intermediate representations, and relation-based approaches~\citep{tung2019similarity,huang2022knowledge,li2022knowledge}, which transfer structural relationships among samples or features. Our proposed method falls under logit-based category, as it extends the standard KD~\citep{hinton2015distilling} by applying an adaptive temperature mechanism for both teacher and student.

\boldhdr{Temperature scaling in KD}
Temperature $\tau$ is a central mechanism in logit-based KD. By dividing logits by a temperature $\tau$ before softmax, KD smooths the teacher distribution and reveals non-target class information that is absent from hard labels~\citep{hinton2015distilling,tang2020understanding,liu2022meta,zhao2022decoupled}. Most existing methods, however, use a fixed temperature for all samples and often share the same temperature between teacher and student. This design is convenient but sample-agnostic: it assumes that a single smoothing strength is suitable for all teacher predictions, despite large variations in logit scale, and sample difficulty.

Recent studies have begun to revisit the role of temperature in KD, but they address different aspects of the problem. CTKD~\citep{li2023curriculum} learns a curriculum-based temperature to control distillation difficulty, but still uses a shared teacher--student temperature. \citet{chandrasegaran2022revisiting} study how smoothing interacts with teacher quality and label smoothing, showing that inappropriate smoothing can weaken distillation. Logit Standardization~\citep{sun2024logit} normalizes logits to reduce sensitivity to temperature choice, but retains a fixed shared base temperature for stability. \citet{su2025ea} use entropy for sample-wise loss reweighting but still rely on the single global temperature of standard KD, whereas our work uses entropy to regulate the informativeness of teacher soft labels.
In contrast, we address a distinct limitation of fixed teacher temperature: the induced inconsistency in soft-label entropy across samples, which leads to uneven and sometimes uninformative distillation supervision.


\section{Limitations of Fixed and Shared Temperatures}
\label{sec:motivation}

\boldhdr{Preliminaries}
Temperature is used in KD to soften teacher and student output distributions, revealing inter-class relationships. Let $C \in \mathbb{N}$ be the number of classes and $\tau>0$ be the temperature. For the $i$-th sample with teacher logits $\mathbf{v}_i \in \mathbb{R}^{C}$ and student logits $\mathbf{z}_i \in \mathbb{R}^{C}$, their softened distributions are
\begin{equation}
    \mathbf{ p }_i^\tau = \text{softmax}\left(\frac{ \mathbf{ v }_i}{\tau}\right), \quad
    \mathbf{ q }_i^\tau = \text{softmax}\left(\frac{ \mathbf{ z }_i}{\tau}\right).
\end{equation}
The softmax function is defined for class $j$-th as $\text{softmax}(v_{i, j}) = \exp (v_{i, j}) / \sum^{C}_{c=1} \exp( v_{i, c} )$. To quantify the smoothness and informativeness of the teacher's soft output, we compute the entropy of the softened distribution for training sample $i$ as follows:
\begin{equation}
\label{eq:entropy}
    H\left( \mathbf{ p }^{\tau}_i \right) = -\sum_{c=1}^{C} p^{\tau}_{i,c} \log p^{\tau}_{i,c}.
\end{equation}
An effective temperature in KD should produce output distributions with moderate entropy, which is soft enough to reveal inter-class relationships but not so uncertain that the information becomes noisy. Prior work~\citep{hinton2015distilling, tang2020understanding, chandrasegaran2022revisiting} supports this view, showing that well-tuned temperatures help the teacher reveal meaningful inter-class relationships and improve student performance. This highlights the critical role of temperature selection in KD.

\begin{figure}[t]
    \centering
    \begin{subfigure}[b]{0.438\linewidth}
        \centering
        \includegraphics[width=\linewidth]{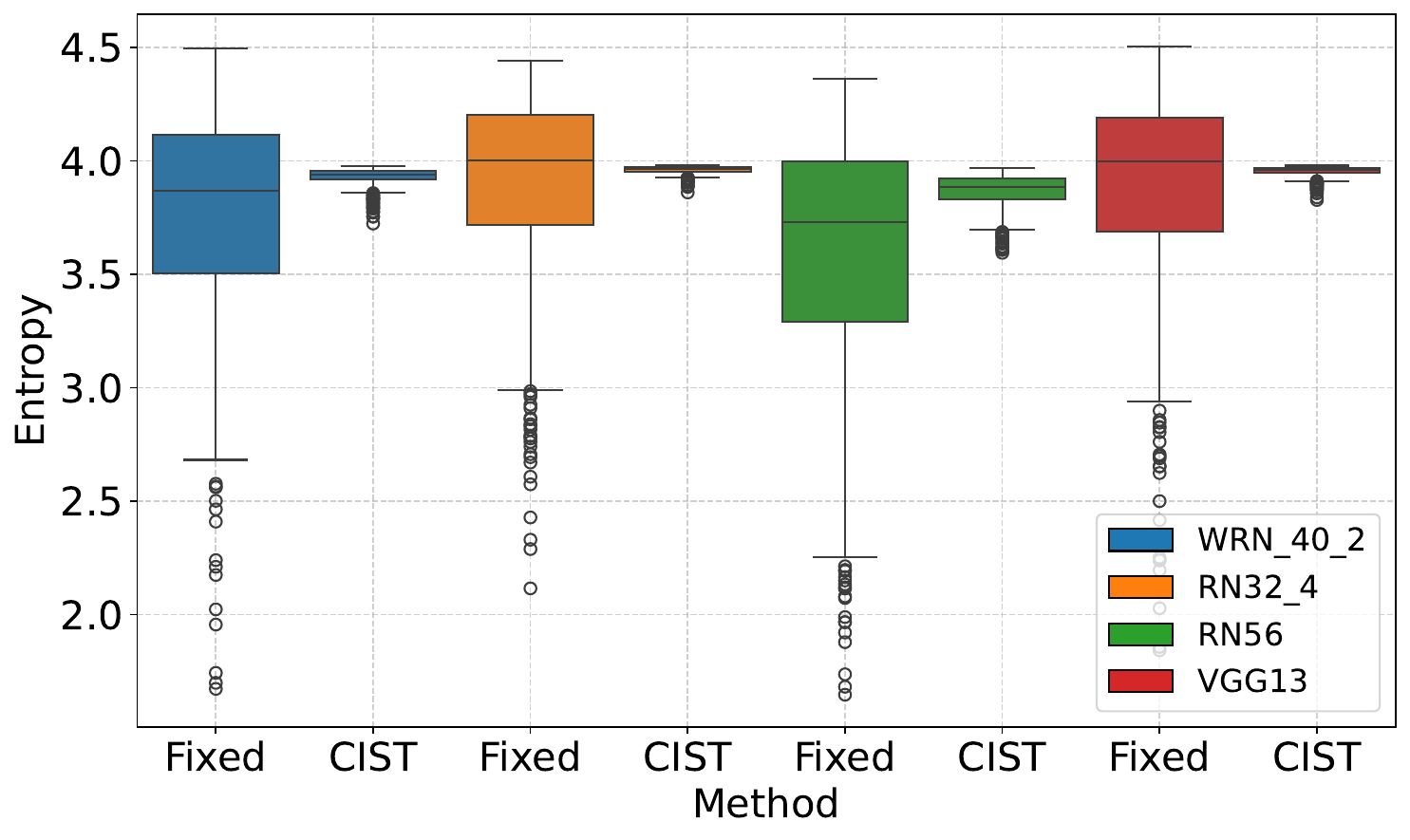}
        \caption{CIFAR-100.}
        \label{fig:compare_cifar}
    \end{subfigure}
    \hfill
    \begin{subfigure}[b]{0.26\linewidth}
        \centering
        \includegraphics[width=\linewidth]{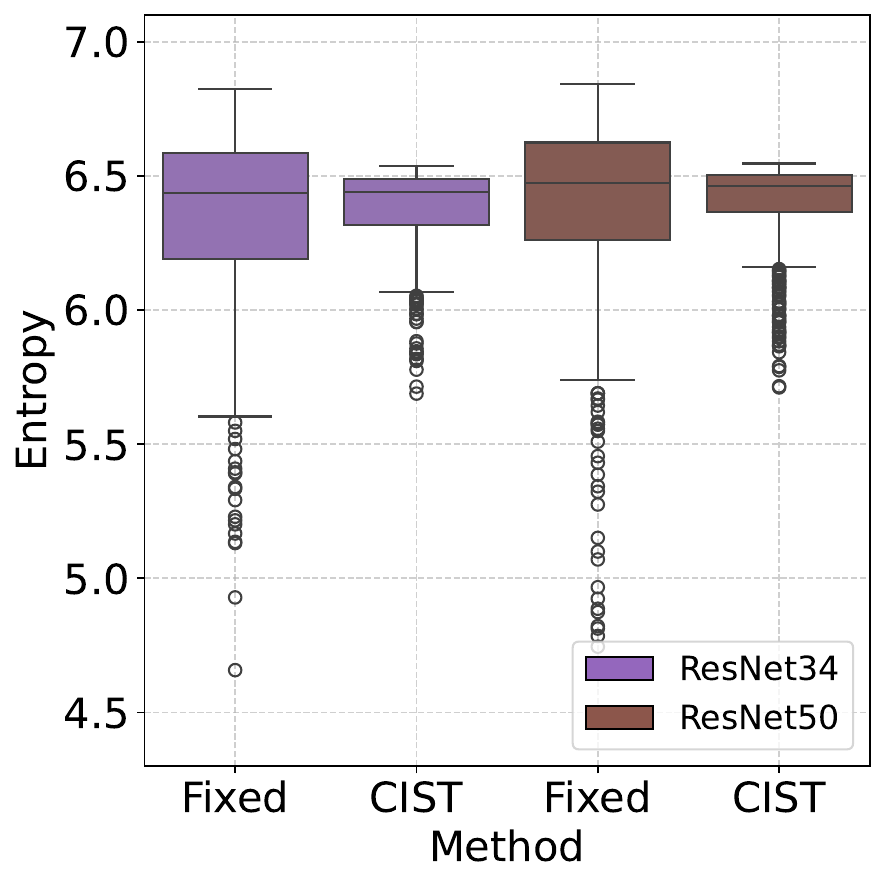}
        \caption{ImageNet.}
        \label{fig:compare_imagenet}
    \end{subfigure}
    \begin{subfigure}[b]{0.26\linewidth}
        \centering
        \includegraphics[width=\linewidth]{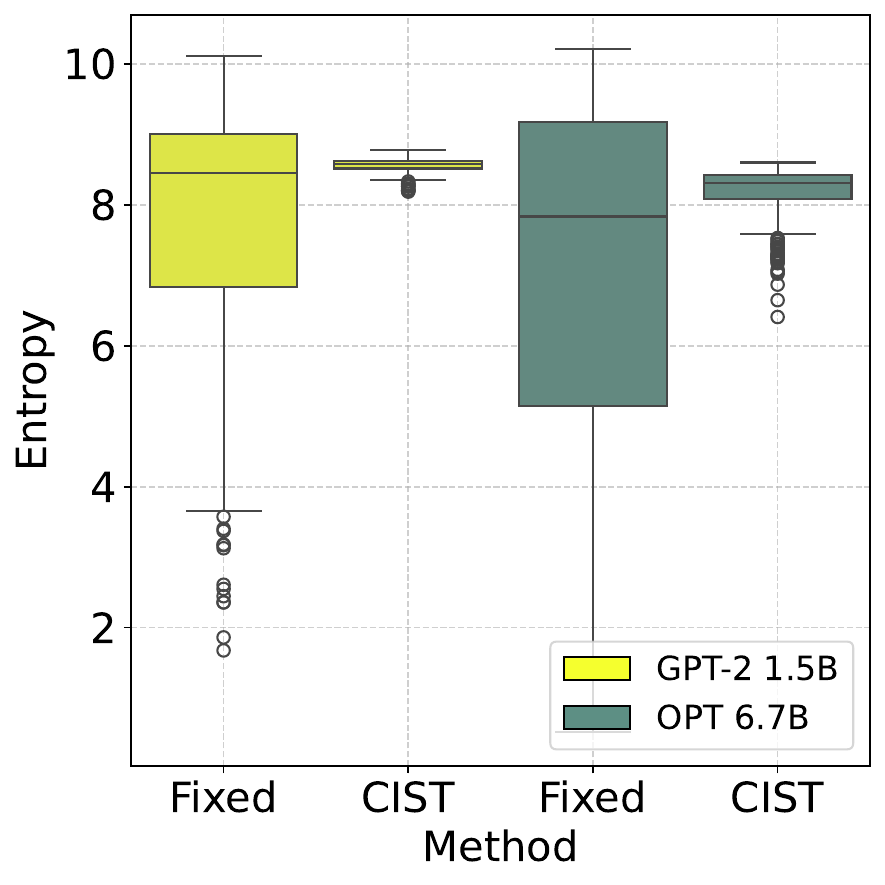}
        \caption{Dolly.}
        \label{fig:compare_dolly}
    \end{subfigure}
    \caption{Entropy distribution of teacher soft labels under standard KD and CIST across architectures and datasets. Soft labels from fixed-temperature KD show high entropy variance and many outliers, indicating ineffective smoothing. In contrast, CIST achieves consistent smoothing.}
    \label{fig:compare}
    \vspace{-0.4cm}
\end{figure}

\boldhdr{Entropy inconsistency under fixed temperature} 
In real-world scenarios, the magnitudes of teacher logits vary substantially due to factors such as sample complexity, teacher confidence, and model inductive biases. As a result, applying a single fixed temperature can produce softened teacher distributions with highly inconsistent entropy.

We illustrate this phenomenon in Figure~\ref{fig:compare}, which shows the entropy distribution of teacher soft labels from CIFAR-100~\citep{krizhevsky2009learning}, ImageNet~\citep{russakovsky2015imagenet}, and Dolly~\citep{DatabricksBlog2023DollyV2}, using various teacher architectures from vision and language models. The results show substantial variation in entropy across samples: some have low entropy (near or even below 2.0 for CIFAR-100), while others reach up to 4.5 for CIFAR-100 and 6.8 for ImageNet, which are close to the entropy of a uniform distribution over 100 classes ($\log(100) \approx 4.6$) and 1000 classes ($\log(1000) \approx 6.9$), respectively. This inconsistency suggests that some soft labels lack meaningful inter-class information, while others are excessively uncertain and resemble random noise. Such variation highlights the limitation of fixed-temperature KD in providing consistently informative supervision, which impairs the effectiveness of distillation.

\begin{wraptable}{r}{0.5\textwidth}
\centering
\scriptsize
\tabcolsep=3pt
\renewcommand\arraystretch{0.9}
\vspace{-0.4cm}
\caption{Test accuracy (\%) on CIFAR-100. Results are averaged over three trials.}
\vspace{-0.2cm}
\label{tab:entropy_outlier}
\scalebox{1.15}{
\begin{tabular}{ccccc}
\toprule
\multirow{2}{*}{Method}  
    & VGG13 & WRN40-2 & RN32x4 & RN56 \\
    & VGG8  & WRN16-2 & RN8x4  & RN20 \\
\midrule
KD       & 72.98 & 74.92 & 73.33 & 70.66 \\
KD-EntOut-CE & \textbf{73.75} & 75.00 & \textbf{74.20} & \textbf{71.05} \\
KD-EntOut-HT & 73.72 & \textbf{75.14} & 73.76 & 70.92 \\
\bottomrule
\vspace{-0.4cm}
\end{tabular}
}
\end{wraptable}

\boldhdr{Impact of entropy outliers} 
To investigate the effect of low-entropy outliers on student performance, we apply different treatments to these samples and examine whether they benefit distillation. Specifically, we define low-entropy outliers as the bottom 5\% of training samples according to the entropy of the teacher soft targets. These samples correspond to highly confident teacher predictions that provide limited inter-class information. We introduce two KD variants:
\begin{itemize}[leftmargin=20pt]
    \item \textit{KD-EntOut-CE}: for entropy outliers, we replace the distillation loss with standard cross-entropy, while keeping distillation unchanged for all other samples.
    \item \textit{KD-EntOut-HT}: we retain distillation for all samples, but use a higher temperature for entropy outliers, increasing $\tau$ from $4$ to $5$ to further soften their predictions.
\end{itemize}
As shown in Table~\ref{tab:entropy_outlier}, both KD-EntOut-CE and KD-EntOut-HT consistently improve over standard KD. For example, KD-EntOut-CE improves VGG13$\rightarrow$VGG8 from $72.98\%$ to $73.75\%$ and RN32x4$\rightarrow$RN8x4 from $73.33\%$ to $74.20\%$, while KD-EntOut-HT achieves the best result on WRN40-2$\rightarrow$WRN16-2. These results suggest that inappropriate smoothing of highly confident teacher predictions can degrade student performance. Moreover, these results indicate that entropy outliers benefit from sample-specific treatment, which can improve student performance. This motivates an effective temperature-scaling strategy for producing consistently informative soft targets.

 \boldhdr{Shared temperature enforces rigid logit matching}
Following the prior analysis~\citep{hinton2015distilling}, we show that using a shared temperature between the teacher and student leads to a rigid logit matching behavior. The distillation loss is computed as the Kullback–Leibler divergence between the softened output probabilities of the student and teacher, denoted by $ q_c $ and $ p_c $, respectively. These are obtained by applying softmax with temperatures $ \tau^s $ (student) and $ \tau^t $ (teacher) to their logits. The gradient of the KD loss with respect to the student logit $z_c$ is:
\begin{align}
    \frac{ \partial L_{\text{KD}} }{ \partial z_c } 
    = \frac{1}{\tau^s} (q_c - p_c)
    = \frac{1}{\tau^s} \left( \frac{ \exp(z_c / \tau^s) }{ \sum_j \exp(z_j / \tau^s) } 
    - \frac{ \exp(v_c / \tau^t) }{ \sum_j \exp(v_c / \tau^t) } \right).
\end{align}
In the high-temperature regime (i.e., when the logits are small relative to the temperature), we apply a first-order Taylor expansion to the exponential and assume zero-mean logits per sample $(\sum_j z_j = \sum_j v_j = 0)$. Under these assumptions, the gradient of the distillation loss simplifies to:
\begin{align}
    \frac{ \partial L_{\text{KD}} }{ \partial z_c } 
    \approx \frac{1}{\tau^s} \left( \frac{ 1 + z_c / \tau^s }{ C + \sum_j z_j / \tau^s }
    - \frac{ 1 + v_c / \tau^t }{ C + \sum_j v_j / \tau^t } \right)
    = \frac{1}{C \tau^s \tau^t} \left( z_c \frac{\tau^t}{\tau^s} - v_c \right).
    \label{eq:ats_relative}
\end{align}
When the teacher and student share the same temperature, $\tau^t=\tau^s=\tau$, this reduces to $1/(C\tau^2) * (z_c - v_c)$ which drives the student toward exact teacher-logit matching. This rigid constraint can be suboptimal, particularly when the teacher and student differ in representational capacity or logit scale~\citep{li2022knowledge,sun2024logit}. 

\section{CIST (Consistently Informative Soft-label Temperature)}
\label{sec:method}

We propose CIST, which consists of three components: (i) sample-wise temperature scaling to stabilize soft-label entropy, (ii) independent teacher--student temperatures to relax rigid logit-scale matching, and (iii) confidence-aware loss rescaling to induce a curriculum over training samples. The pseudo-code for CIST is presented in Algorithm~\ref{algo:at}.

\begin{Proposition}\label{prop:ats}
    Let $\mathbf{v}_i, \mathbf{v}_j \in \mathbb{R}^C$ be the logits for two samples and let $m=\arg\max_c v_{i,c}$ and $m'=\arg\max_c v_{j,c}$. Their softened softmax outputs, computed with temperatures $\tau_i$ and $\tau_j$, are denoted by $\mathbf{p}_i^{\tau_i}$ and $\mathbf{p}_j^{\tau_j}$, respectively. If both distributions are dominated by their maximum logit, the entropy difference $\left| H\left( \mathbf{p}_i^{\tau_i} \right) - H\left( \mathbf{p}_j^{\tau_j} \right) \right|$ is minimized when the ratio of the dominant logit to temperature is equal across samples, i.e., $\frac{v_{i,m}}{\tau_i} = \frac{v_{j,m'}}{\tau_j} = \rho$ for some constant $\rho > 0$.
\end{Proposition}

The proof of Proposition~\ref{prop:ats} is provided in Appendix~\ref{appendix:proof}, which shows that, for a strong well-trained teacher, teacher-label entropy is mainly controlled by the ratio between the dominant teacher logit and the temperature. Thus, a fixed global temperature cannot ensure consistent entropy: high-confidence samples require stronger smoothing, while lower-confidence samples require weaker smoothing. This motivates sample-wise temperature adaptation to match the sample-specific logit scale.

\boldhdr{Sample-wise adaptive temperature}
Following Proposition~\ref{prop:ats}, CIST chooses the teacher temperature by normalizing the dominant logit to a constant $\rho$:
\begin{equation}
    \tau^t_i = \frac{v_{i, m}}{\rho}.
\end{equation} 
This formulation assigns the larger temperatures to highly confident teacher predictions, preventing them from remaining overly sharp, while assigning smaller temperatures to less confident predictions, preventing excessive smoothing. As a result, CIST reduces entropy variation across samples and produces teacher soft labels with more consistent informativeness.

To empirically evaluate entropy stabilization, we compare the entropy distributions of teacher soft labels produced by fixed-temperature KD and CIST across different teacher architectures and datasets. As shown in Figure~\ref{fig:compare}, fixed-temperature KD exhibits high entropy variance, with many samples producing either overly sharp or overly smooth targets. This pattern is consistent across both vision and language datasets. Note that the entropy ranges differ across datasets due to the different numbers of classes. In contrast, CIST substantially reduces entropy variance across all three datasets and produces more consistent soft-label entropy, with values concentrated around $5.75$--$6.5$ on ImageNet.

\boldhdr{Independent teacher and student temperature}
Standard KD typically uses the same temperature for the teacher and student, implicitly forcing them to align under an exact logit scale despite their capacity mismatch. CIST addresses this issue by assigning separate sample-wise temperatures to the teacher and student. Let $v_{i,m}=\max_c v_{i,c}$ denote the dominant teacher logit, and let $z_{i,m'}=\max_c |z_{i,c}|$ denote the maximum student logit magnitude for sample $i$. We define the teacher and student temperatures as
$
    \tau_i^t = \frac{v_{i,m}}{\rho},
    \
    \tau_i^s = \frac{z_{i,m'}}{\rho},
$
respectively.
Using separate temperatures allows the teacher and student predictions to be smoothed according to their respective logit scales. Substituting these temperatures into the high-temperature approximation in Eq.~\eqref{eq:ats_relative} gives
\begin{equation}
\label{eq:grad_ats}
    \frac{\partial L_{\mathrm{KD}}}{\partial z_c}
    \approx
    \frac{1}{C\tau_i^s\tau_i^t}
    \left(
    z_{i,c}\frac{v_{i,m}}{z_{i,m'}}
    -
    v_{i,c}
    \right).
\end{equation}
CIST introduces the relative scale factor $v_{i,m}/z_{i,m'}$, allowing the student to match the teacher's logit structure without being forced to reproduce the teacher's absolute logit magnitude. This relaxes rigid logit-scale matching and encourages the student to preserve inter-class relations.

\boldhdr{Confidence-aware curriculum regularization}
Adaptive temperatures change both the shape of the softened distributions and the scale of the KD gradients. As shown in Eq.~\eqref{eq:grad_ats}, the gradient magnitude is inversely proportional to $\tau_i^s\tau_i^t$. Therefore, to maintain a stable balance between the KL distillation loss and the hard-label cross-entropy loss, we follow the standard temperature-scaled KD principle~\citep{hinton2015distilling} and rescale the KL term for each sample $i$:
\begin{equation}
    \mathcal{L}_{\text{KL}} = \tau_i^t \tau_i^s \cdot \mathrm{KL} \left( \mathbf{p}_i^{\tau_i^t} \parallel \mathbf{q}_i^{\tau_i^s} \right),
\end{equation}
where $ \mathbf{p}_i^{\tau^t} $ and $ \mathbf{q}_i^{\tau^s} $ are the temperature-scaled teacher and student output probabilities. Importantly, this rescaling factor also acts as a confidence-aware curriculum weight. Since $\tau_i^t$ and $\tau_i^s$ are determined by the teacher and student logit magnitudes, their product assigns larger weights to samples where the teacher provides confident soft targets and the student produces confident responses. Conversely, samples with low teacher confidence or weak student responses receive smaller weights, reducing the influence of unreliable or difficult supervision. Thus, CIST induces a sample-wise curriculum: it prioritizes distillation signals that are both reliable from the teacher and learnable for the student.

\section{Experiments}

\subsection{Implementation details}

\boldhdr{Vision task} 
We evaluate on CIFAR-100~\citep{krizhevsky2009learning} and ImageNet~\citep{russakovsky2015imagenet}. Following standard KD protocols~\citep{tian2019contrastive,zhao2022decoupled}, we consider diverse teacher--student architectures, including VGG~\citep{simonyan2014very}, ResNet (RN)~\citep{he2016deep}, WideResNet (WRN)~\citep{zagoruyko2016wide}, MobileNet (MN)~\citep{howard2017mobilenets,sandler2018mobilenetv2}, and ShuffleNet (SHN)~\citep{zhang2018shufflenet}. We compare against logit-based KD methods, including KD~\citep{hinton2015distilling}, DKD~\citep{zhao2022decoupled}, CTKD~\citep{li2023curriculum}, MLKD~\citep{jin2023multi}, LS~\citep{sun2024logit}, and EA~\citep{su2025ea}, as well as feature-based methods including FitNet~\citep{romero15-iclr}, RKD~\citep{park2019relational}, CRD~\citep{peng2019correlation}, OFD~\citep{heo2019comprehensive}, and ReviewKD~\citep{chen2021distilling}.

\boldhdr{Language task} 
We use the instruction--response data from \texttt{databricks-dolly-15k}~\citep{DatabricksBlog2023DollyV2} for training. The dataset contains 14k training samples, 500 validation samples, and 500 test samples. We evaluate distillation from GPT-2 XL~\citep{radford2019language} and OPT-6.7B~\citep{zhang2022opt} teachers to GPT-2 Base and OPT-1.3B students. Evaluation is conducted on four benchmarks consisting of Dolly Eval~\citep{DatabricksBlog2023DollyV2}, Vicuna Eval~\citep{vicuna2023}, Super-Natural Instructions~\citep{wang2022super}, and Unnatural Instructions~\citep{honovich2023unnatural}. We compare CIST with common KL-based baselines in language model distillation, including FKL, RKL~\citep{gu2023minillm}, Sym-KL, JS~\citep{agarwal2024policy}, SRKL~\citep{ko2024distillm}, AKL~\citep{wu2025rethinking}, and $\alpha$-$\beta$ divergence (AB)~\citep{wang2025abkd}. 

Additional implementation details for both tasks, including training resources and hyperparameter settings, are provided in the Appendix~\ref{appendix:implementationdetails}.

\subsection{Results on Vision Task}

\newcommand{\imp}[1]{\textcolor{red}{{\fontsize{4}{4}\selectfont #1}}}
\newcommand{\impbox}[1]{\hspace{1pt}\makebox[1.6em][l]{\imp{#1}}}

\definecolor{deltagreen}{RGB}{0,120,60}
\definecolor{oursblue}{RGB}{232,242,255}

\newcommand{\deltarow}[1]{\textcolor{deltagreen}{\scriptsize #1}}

\begin{table*}[tbp]
		\centering
		\scriptsize
        \caption{Top-1 test accuracy (\%) on CIFAR-100. Distillation methods are grouped by category, and student-teacher pairs are grouped by architecture type. The best and second-best results are marked in \textbf{bold} and \underline{underlined}, respectively. $\Delta_{\text{KD}}$ and $\Delta_{\text{MLKD}}$ report the improvement of KD+Ours over KD and MLKD+Ours over MLKD, respectively. The results are averaged over three trials.}
		\setlength\tabcolsep{4pt}
		\renewcommand\arraystretch{1}
		\scalebox{.97}{
			\begin{tabular}{ll|ccccc|ccccc}
				\hline
				& & \multicolumn{5}{c}{\textbf{Homogeneous}} & \multicolumn{5}{c}{\textbf{Heterogeneous}} \\
				\hline
				\multirow{4}*{\textbf{Type}}
                        & \multirow{2}{*}{Teacher} & \makebox[0.06\textwidth][c]{WRN40-2} & \makebox[0.06\textwidth][c]{RN56} & \makebox[0.06\textwidth][c]{RN110} & \makebox[0.06\textwidth][c]{RN32x4} & \makebox[0.06\textwidth][c]{VGG13} & \makebox[0.06\textwidth][c]{RN32x4} & \makebox[0.06\textwidth][c]{RN32×4} & \makebox[0.06\textwidth][c]{WRN40-2} & \makebox[0.06\textwidth][c]{RN50} & \makebox[0.06\textwidth][c]{RN32x4} \\
    				& & 75.61 & 72.34 & 74.31 & 79.42 & 74.64 & 79.42 & 79.42 & 75.61 & 79.34 & 79.42 \\
    				& \multirow{2}{*}{Student} & WRN16-2 & RN20 & RN32 & RN8x4 & VGG8 & WRN16-2 & WRN40-2 & RN8x4 & MN-V2 & SHN-V2 \\
    				& & 73.26 & 69.06 & 71.14 & 72.50 & 70.36 & 73.26 & 75.61 & 72.50 & 64.60 & 71.82 \\
				\midrule
                    \multirow{6}{*}{\textbf{Feat}}
                        & FitNet    & 73.58 & 69.21 & 71.06 & 73.50 & 71.02 & 74.70 & 77.69 & 74.61 & 63.16 & 73.54 \\
                        & AT        & 74.08 & 70.55 & 72.31 & 73.44 & 71.43 & 73.91 & 77.43 & 74.11 & 58.58 & 72.73 \\
                        & RKD       & 73.35 & 69.61 & 71.82 & 71.90 & 71.48 & 74.86 & 77.82 & 75.26 & 64.43 & 73.21 \\
                        & CRD       & 75.48 & 71.16 & 73.48 & 75.51 & 73.94 & 75.65 & 78.15 & 75.24 & 69.11 & 75.65 \\
                        & OFD       & 75.24 & 70.98 & 73.23 & 74.95 & 73.95 & 76.17 & 79.25 & 74.36 & 69.04 & 76.82 \\
                        & ReviewKD  & 76.12 & 71.89 & 73.89 & 75.63 & 74.84 & 76.11 & 78.96 & 74.34 & 69.89 & 77.78 \\
                    \midrule
				\multirow{10}*{\textbf{Logit}}
    				& KD        & 74.92 & 70.66 & 73.08 & 73.33 & 72.98 & 74.90 & 77.70 & 73.97 & 67.35 & 74.45 \\
                        & CTKD       & 75.45 & 71.19 & 73.52 & 73.39 & 73.52 & 74.57 & 77.66 & 74.61 & 68.67 & 75.37 \\
                        & DKD       & 76.24 & 71.97 & 74.11 & 76.32 & 74.68 & 75.70 & 78.46 & 75.56 & 70.35 & 77.07 \\
                        & MLKD   & \ul{76.63} & \ul{72.19} & 74.11 & \ul{77.08} & \ul{75.18} & \ul{76.52} & \ul{79.26} & 77.33 & \ul{71.04} & \ul{78.44} \\
                        & KD+EA   & 75.49 & 71.60 & 73.66 & 75.46 & 74.08 & 75.08 & 77.59 & 74.84 & 69.67 & 75.91 \\
                        & KD+LS   & 76.11 & 71.43 & 74.17 & 76.62 & 74.36 & 75.26 & 77.92 & 77.11 & 69.02 & 75.56 \\
                        & \cellcolor{oursblue}KD+Ours   & \cellcolor{oursblue}76.59 & \cellcolor{oursblue}71.97 & \cellcolor{oursblue}\ul{74.30} & \cellcolor{oursblue}76.96 & \cellcolor{oursblue}74.82 
                                    & \cellcolor{oursblue}76.32 & \cellcolor{oursblue}78.82 & \cellcolor{oursblue}\ul{77.36} & \cellcolor{oursblue}69.87 & \cellcolor{oursblue}76.60 \\
                        & $\Delta_{\text{KD}}$ 
                                    & \deltarow{+1.67} & \deltarow{+1.31} & \deltarow{+1.22} & \deltarow{+3.63} & \deltarow{+1.84} 
                                    & \deltarow{+1.42} & \deltarow{+1.12} & \deltarow{+3.39} & \deltarow{+2.52} & \deltarow{+2.15} \\

                        & \cellcolor{oursblue}MLKD+Ours & \cellcolor{oursblue}\textbf{77.12} & \cellcolor{oursblue}\textbf{72.52} & \cellcolor{oursblue}\textbf{74.62} & \cellcolor{oursblue}\textbf{78.71} & \cellcolor{oursblue}\textbf{75.50} 
                                    & \cellcolor{oursblue}\textbf{77.69} & \cellcolor{oursblue}\textbf{79.79} & \cellcolor{oursblue}\textbf{78.54} & \cellcolor{oursblue}\textbf{71.33} & \cellcolor{oursblue}\textbf{78.92} \\
                        & $\Delta_{\text{MLKD}}$ 
                            & \deltarow{+0.49} & \deltarow{+0.33} & \deltarow{+0.51} & \deltarow{+1.63} & \deltarow{+0.32} 
                            & \deltarow{+1.17} & \deltarow{+0.53} & \deltarow{+1.21} & \deltarow{+0.29} & \deltarow{+0.48} \\

                    \bottomrule
			\end{tabular}
		}
    \label{table:same_arch}
    \vspace{-0.3cm}
\end{table*}

\boldhdr{CIFAR-100}
Table~\ref{table:same_arch} reports Top-1 accuracy on CIFAR-100 across homogeneous (same architecture family) and heterogeneous (different architecture family) teacher--student pairs. Overall, MLKD+Ours consistently achieves the best performance, surpassing the second-best method by $0.29\%$--$1.63\%$. These gains are significant in KD, where improvements above $0.15\%$ are considered non-trivial in recent work~\citep{sun2024logit} and our smallest gain is nearly twice this threshold.

In the \emph{homogeneous setting}, KD+Ours substantially improves standard KD, with gains of $+3.63\%$ on RN32x4$\rightarrow$RN8x4 and $+1.67\%$ on WRN40-2$\rightarrow$WRN16-2. It also outperforms strong logit-based baselines such as DKD, while remaining competitive with feature-based methods such as ReviewKD. When combined with MLKD, our method further improves performance by $0.30\%$--$1.63\%$ over MLKD and achieves the best results across all homogeneous pairs, including a gain exceeding $5\%$ over standard KD on RN32x4$\rightarrow$RN8x4.

In the \emph{heterogeneous setting}, where architectural mismatch makes distillation more challenging, KD+Ours outperforms standard KD and other logit-based baselines in most cases. Although feature-based methods such as OFD and ReviewKD benefit from explicit intermediate-representation alignment, MLKD+Ours consistently achieves the best overall performance. It surpasses the second-best method by up to $1.18\%$ on WRN40-2$\rightarrow$ResNet8x4 and outperforms feature-based baselines by margins ranging from $0.5\%$ to over $3\%$.

\definecolor{oursblue}{RGB}{232,242,255}

\begin{wraptable}{r}{0.45\textwidth}
\vspace{-4mm}
\centering
\footnotesize
\tabcolsep=2.5pt
\renewcommand\arraystretch{0.95}
\caption{Top-1 and top-5 accuracy on the ImageNet validation set. The best and second best results are in \textbf{bold} and \underline{underlined}.}
\resizebox{\linewidth}{!}{
\begin{tabular}{lcccc}
\toprule
Teacher/Student 
& \multicolumn{2}{c}{ResNet34/ResNet18} 
& \multicolumn{2}{c}{ResNet50/MN-V1} \\
\cmidrule(lr){2-3}\cmidrule(lr){4-5}
Accuracy  & top-1 & top-5 & top-1 & top-5 \\
\midrule
Teacher & 73.31 & 91.42 & 76.16 & 92.86 \\
Student & 69.75 & 89.07 & 68.87 & 88.76 \\
\midrule
AT       & 70.69 & 90.01 & 69.56 & 89.33 \\
CRD      & 71.17 & 90.13 & 71.37 & 90.41 \\
OFD      & 70.81 & 89.98 & 71.25 & 90.34 \\
ReviewKD & 71.61 & \underline{90.51} & \underline{72.56} & 91.00 \\
\midrule
KD       & 71.03 & 90.05 & 70.50 & 89.80 \\
CTKD     & 71.38 & 90.27 & 71.16 & 90.11 \\
DKD      & 71.70 & 90.41 & 72.05 & \underline{91.05} \\
KD+LS    & 71.42 & 90.29 & 72.18 & 90.80 \\
\cellcolor{oursblue}KD+Ours  & \cellcolor{oursblue}\textbf{71.97} & \cellcolor{oursblue}\textbf{90.73} & \cellcolor{oursblue}\textbf{72.75} & \cellcolor{oursblue}\textbf{91.41} \\
\bottomrule
\end{tabular}
}
\label{tab:imagenet}
\vspace{-3mm}
\end{wraptable}

\boldhdr{ImageNet} 
Table~\ref{tab:imagenet} compares CIST (KD+Ours) with representative feature-based and logit-based distillation methods on ImageNet in terms of Top-1 and Top-5 accuracy. For a fair efficiency comparison, we exclude MLKD since it incurs substantially higher training cost than standard KD, whereas CIST preserves the training cost of logit-based KD. Overall, KD+Ours achieves the best performance among all compared methods across both logit- and feature-based categories. 
In particular, KD+Ours consistently surpasses the strongest logit-based baselines, including DKD and KD+LS, by clear margins: +0.27\% on ResNet34-ResNet18 and +0.57\% on ResNet50-MobileNetV1. Compared to standard KD, CIST yields up to +2.25\% Top-1 improvement on ResNet50-MobileNetV1, demonstrating its effectiveness under challenging teacher--student capacity gaps. Moreover, KD+Ours also outperforms strong feature-based methods such as OFD and ReviewKD, which require additional trainable modules and introduce non-trivial training overhead.

\begin{figure}[t]
    \centering
    \begin{subfigure}[b]{0.29\linewidth}
        \centering
        \includegraphics[width=\linewidth]{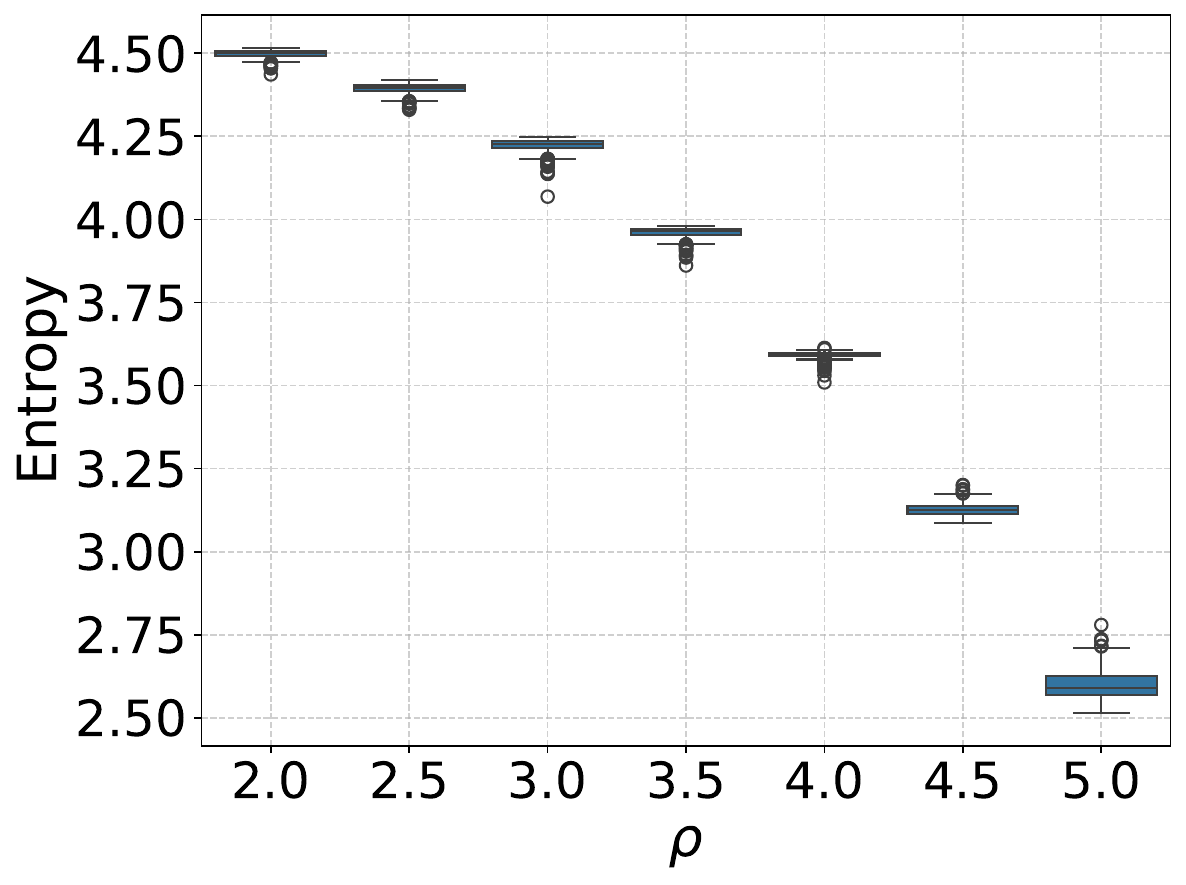}
        \caption{}
        \label{fig:entropy_rho}
    \end{subfigure}
    \hfill
    \begin{subfigure}[b]{0.295\linewidth}
        \centering
        \includegraphics[width=\linewidth]{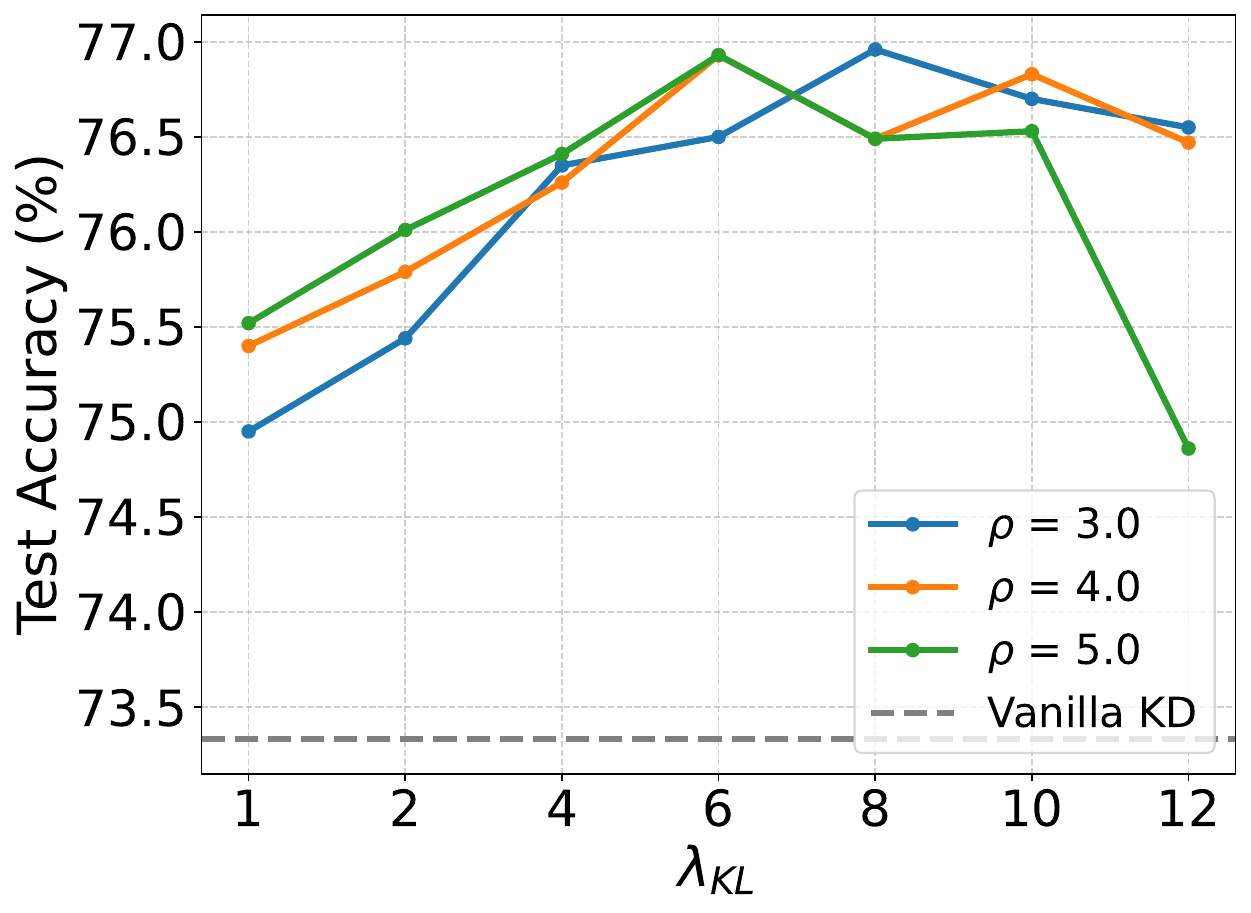}
        \caption{}
        \label{fig:alpha_analysis}
    \end{subfigure}
    \hfill
    \begin{subfigure}[b]{0.38\linewidth}
        \centering
        \includegraphics[width=\linewidth]{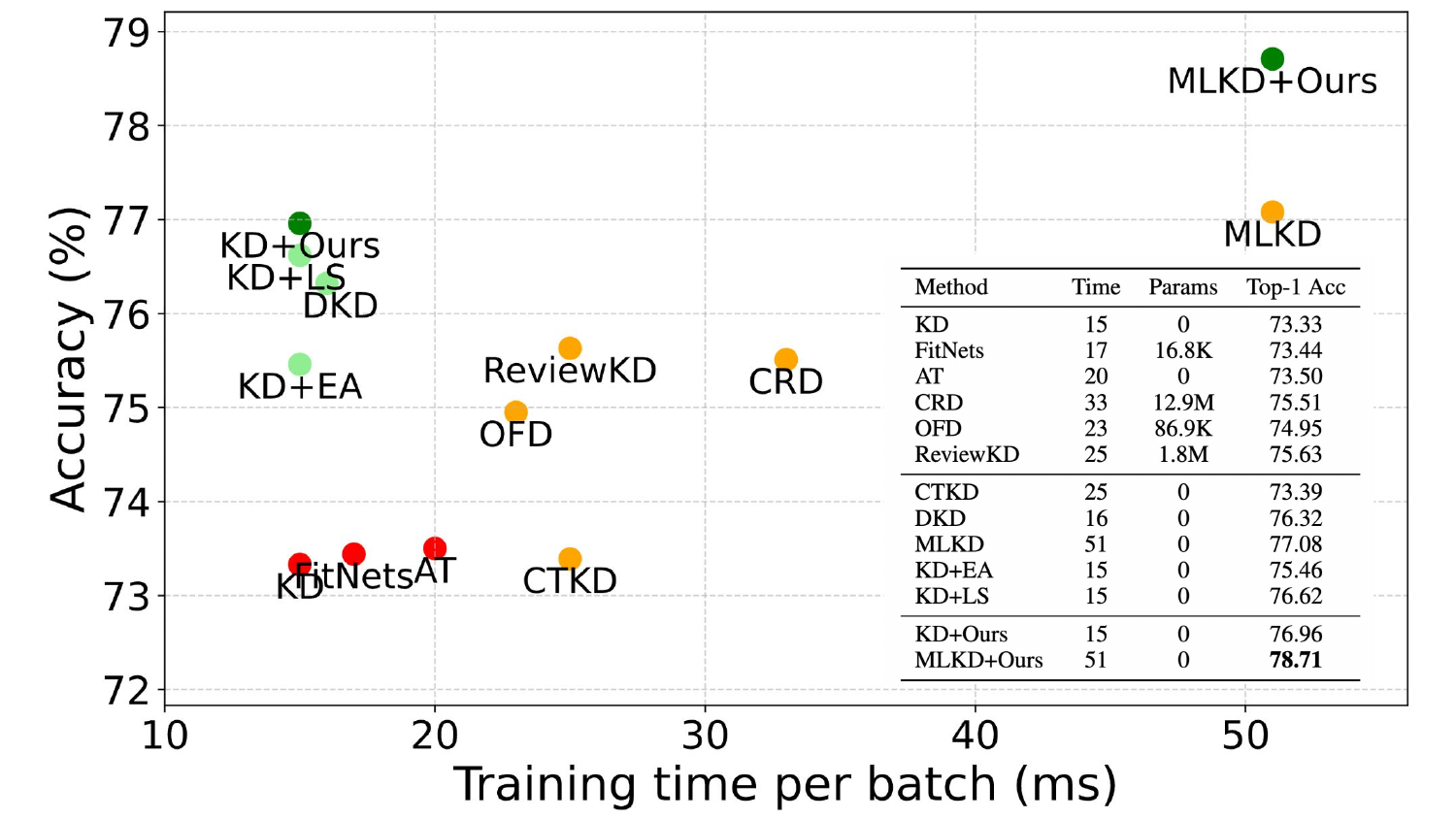}
        \caption{}
        \label{fig:training_efficiency}
    \end{subfigure}
    \caption{(a) Effect of $\rho$ on the entropy distribution of teacher soft targets from ResNet32x4. (b) The impact of various $\lambda_{\text{KL}}$ and $\rho$ values on student test accuracy (\%) in CIST. (c) Training time per batch and test accuracy (\%) of different distillation methods on the ResNet32x4-ResNet8x4 pair.
}
\end{figure}

\subsection{Results on Language Task}

\definecolor{teachergray}{RGB}{245,245,245}
\definecolor{oursblue}{RGB}{232,242,255}

\begin{table}[t]
    \centering
    \vspace{-4mm}
    \scriptsize
    \caption{ROUGE-L for GPT-2 and OPT teacher--student pairs, with mean and standard deviation computed over five random seeds. The best results are \textbf{boldfaced} and the second-best are \underline{underlined}.}
    \setlength{\tabcolsep}{10pt}
    \renewcommand{\arraystretch}{1.08}
    \resizebox{\linewidth}{!}{
        \begin{tabular}{llccccc}
        \toprule
        \textbf{Student}
        & \textbf{Method / Teacher}  
        & \textbf{Dolly}  
        & \textbf{Vicuna}  
        & \textbf{Super-NI} 
        & \textbf{UnNI} 
        & \textbf{Avg.} ($\uparrow$)\\
        \midrule

        \multirow{9}{*}{\shortstack[l]{GPT-2\\0.1B}}
        & \cellcolor{teachergray} Teacher: GPT-2 1.5B
        & \cellcolor{teachergray}\rateinline{27.00}{0.19}
        & \cellcolor{teachergray}\rateinline{16.31}{0.32}
        & \cellcolor{teachergray}\rateinline{26.46}{0.41}
        & \cellcolor{teachergray}\rateinline{31.10}{0.06} 
        & \cellcolor{teachergray}25.22 \\

        \cmidrule(lr){2-7}

        & FKL 
        & \rateinline{23.80}{0.55} 
        & \rateinline{14.53}{0.31} 
        & \rateinline{16.27}{0.24} 
        & \rateinline{19.03}{0.09} 
        & 18.41 \\
        
        & RKL 
        & \rateinline{\underline{24.67}}{0.13} 
        & \rateinline{\ul{15.82}}{0.43} 
        & \rateinline{21.03}{0.21} 
        & \rateinline{23.94}{0.12} 
        & 21.37 \\
        
        & Sym-KL 
        & \rateinline{24.38}{0.14} 
        & \rateinline{15.72}{0.75} 
        & \rateinline{19.92}{0.25} 
        & \rateinline{22.21}{0.03} 
        & 20.56 \\
        
        & JS
        & \rateinline{24.08}{0.26} 
        & \rateinline{15.47}{0.37} 
        & \rateinline{19.66}{0.23} 
        & \rateinline{22.17}{0.21} 
        & 20.34 \\
        
        & SRKL
        & \rateinline{24.48}{0.37} 
        & \rateinline{14.96}{0.34} 
        & \rateinline{\textbf{23.25}}{0.36} 
        & \rateinline{24.01}{0.05} 
        & 21.68 \\
        
        & AKL 
        & \rateinline{21.83}{0.19} 
        & \rateinline{13.70}{0.21} 
        & \rateinline{15.40}{0.20} 
        & \rateinline{18.06}{0.13} 
        & 17.25 \\
        
        & AB
        & \rateinline{24.32}{0.29} 
        & \rateinline{\ul{15.82}}{0.58} 
        & \rateinline{\underline{23.08}}{0.19} 
        & \rateinline{\underline{24.32}}{0.14} 
        & \underline{21.88} \\

        & \cellcolor{oursblue}CIST (Ours)  
        & \cellcolor{oursblue}\rateinline{\textbf{25.42}}{0.10}  
        & \cellcolor{oursblue}\rateinline{\textbf{16.10}}{0.07}
        & \cellcolor{oursblue}\rateinline{21.81}{0.23}
        & \cellcolor{oursblue}\rateinline{\textbf{25.94}}{0.13}
        & \cellcolor{oursblue}\textbf{22.32}  \\

        \midrule

        \multirow{9}{*}{\shortstack[l]{OPT\\1.3B}}
        & \cellcolor{teachergray} Teacher: OPT 6.7B
        & \cellcolor{teachergray}\rateinline{27.52}{0.29} 
        & \cellcolor{teachergray}\rateinline{17.64}{0.27}
        & \cellcolor{teachergray}\rateinline{30.41}{0.46}
        & \cellcolor{teachergray}\rateinline{31.39}{0.20} 
        & \cellcolor{teachergray}26.74 \\

        \cmidrule(lr){2-7}
        
        & FKL 
        & \rateinline{26.07}{0.65} 
        & \rateinline{16.71}{0.33} 
        & \rateinline{22.11}{0.38} 
        & \rateinline{\underline{26.82}}{0.10}  
        & 22.93 \\

        & RKL
        & \rateinline{26.58}{0.11} 
        & \rateinline{17.54}{0.13} 
        & \rateinline{22.64}{0.16} 
        & \rateinline{26.29}{0.10}   
        & 23.26 \\

        & Sym-KL 
        & \rateinline{25.73}{0.40} 
        & \rateinline{16.95}{0.37} 
        & \rateinline{23.40}{0.22} 
        & \rateinline{25.44}{0.07}  
        & 22.88 \\

        & JS 
        & \rateinline{26.39}{0.71} 
        & \rateinline{17.15}{0.39} 
        & \rateinline{24.02}{0.31} 
        & \rateinline{26.59}{0.11}  
        & 23.54 \\

        & SRKL
        & \rateinline{26.04}{0.36} 
        & \rateinline{17.34}{0.25} 
        & \rateinline{22.96}{0.22} 
        & \rateinline{24.71}{0.22}  
        & 22.76 \\

        & AKL 
        & \rateinline{26.20}{0.27} 
        & \rateinline{17.26}{0.66} 
        & \rateinline{22.12}{0.18} 
        & \rateinline{25.89}{0.20}  
        & 22.87 \\

        & AB
        & \rateinline{\underline{26.86}}{0.17} 
        & \rateinline{\underline{17.59}}{0.41} 
        & \rateinline{\underline{24.39}}{0.11} 
        & \rateinline{26.62}{0.09}  
        & \underline{23.87} \\

        & \cellcolor{oursblue}CIST (Ours)  
        & \cellcolor{oursblue}\rateinline{\textbf{27.74}}{0.26}  
        & \cellcolor{oursblue}\rateinline{\textbf{17.69}}{0.23}
        & \cellcolor{oursblue}\rateinline{\textbf{25.19}}{0.25}
        & \cellcolor{oursblue}\rateinline{\textbf{28.41}}{0.12}
        & \cellcolor{oursblue}\textbf{24.76}  \\

        \bottomrule
        \end{tabular}
    }
    \label{tab:main1}
    \vspace{-0.2cm}
\end{table}

\boldhdr{GPT-2}
Table~\ref{tab:main1} shows that CIST consistently delivers strong performance across instruction-following benchmarks, ranking first on three out of four tasks and achieving the best average score. On UnNI, CIST yields the largest gain, outperforming RKL by $2.00$ and the second-best method by $1.62$. On Dolly and Vicuna, CIST further improves over the second-best method by $0.75$ and $0.28$, respectively. Although CIST ranks third on Super-NI, slightly behind AB and SRKL, its overall performance highlights its robustness across diverse evaluation tasks.

\boldhdr{OPT}
For larger teacher--student model pairs, CIST consistently outperforms state-of-the-art distillation objectives, ranking first on all four benchmarks and achieving the best average score. On Dolly, CIST exceeds the second-best method, AB, by $0.88$ points and the strong RKL baseline by $1.16$ points. On Vicuna, although the improvement over the second-best method is smaller ($0.10$ points), CIST obtains a lower standard deviation, indicating more stable gains. The largest improvement is observed on UnNI, where CIST outperforms AB by $1.79$ points.

\subsection{Ablation Study}

 \boldhdr{Impact of $\rho$ on entropy and student performance} 
As shown in Figure~\ref{fig:entropy_rho}, increasing $\rho$ leads to lower adaptive temperatures $\tau_i$, producing sharper teacher outputs with reduced entropy. In contrast, smaller $\rho$ values yield larger $\tau_i$ and higher-entropy soft targets. For example, the average entropy decreases from 4.5 (nearly uniform distribution) to 2.6 as $\rho$ increases from 2.0 to 5.0. These findings confirm that the hyperparameter $\rho$ provides effective control over the entropy of the teacher’s soft labels and maintains consistent entropy behavior across diverse samples, even when $\rho$ varies widely.

To evaluate the sensitivity of CIST to the hyperparameter $\rho$, we conduct experiments on the ResNet32x4$\rightarrow$ResNet8x4 pair with $\rho \in \{3.0, 4.0, 5.0\}$ and KL loss weight $\lambda_{\mathrm{KL}} \in \{1, 2, 4, 6, 8, 10, 12\}$. As shown in Figure~\ref{fig:alpha_analysis}, CIST consistently outperforms vanilla KD across all configurations, demonstrating its robustness to the choice of $\rho$. Performance remains stable for $\lambda_{\mathrm{KL}}$ in the range of $4$ to $10$, while overly large distillation weights can degrade accuracy. The best result is achieved with $\rho=3.0$ and $\lambda_{\mathrm{KL}}=8$, reaching $76.96\%$ accuracy.

\boldhdr{t-SNE visualization} 
We present t-SNE visualizations using ResNet32x4 as the teacher and ResNet8x4 as the student on CIFAR-100. The results demonstrate that the representations learned with CIST are more class-separable compared to those from standard KD, indicating that CIST enhances the discriminability of learned features.

\begin{figure}[t]
    \centering
    \begin{subfigure}[b]{0.3\linewidth}
        \centering
        \includegraphics[width=\linewidth]{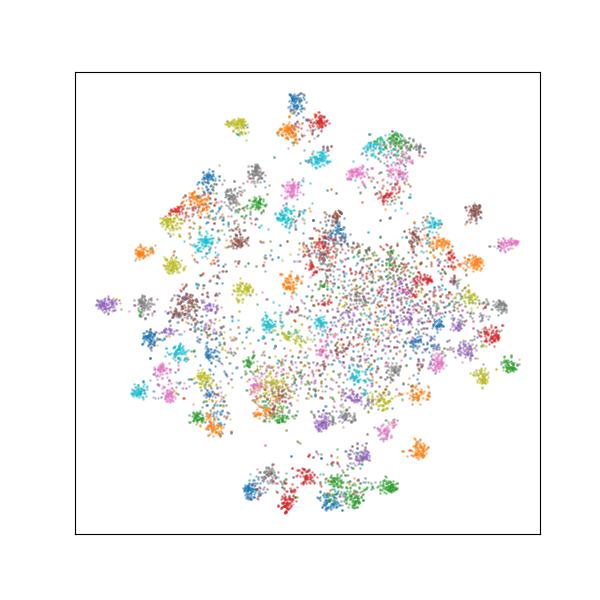}
        \vspace{-0.4cm}
        \caption{KD}
        \label{fig:tsne_kd}
    \end{subfigure}
    \hfill
    \begin{subfigure}[b]{0.3\linewidth}
        \centering
        \includegraphics[width=\linewidth]{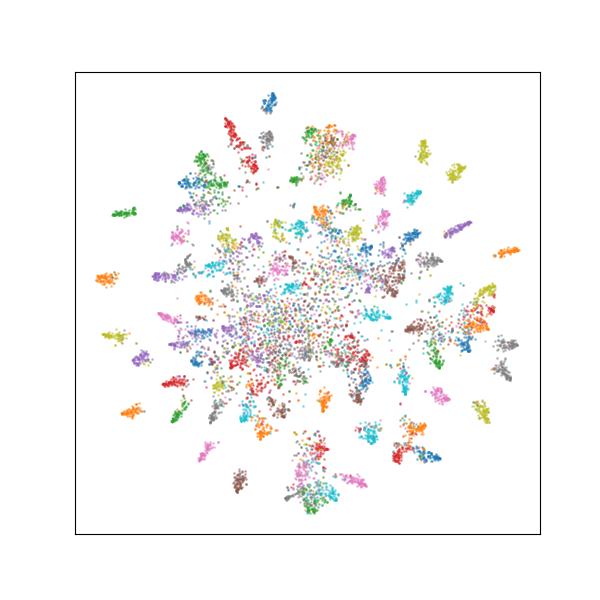}
        \vspace{-0.4cm}
        \caption{CIST}
        \label{fig:tsne_ats}
    \end{subfigure}
    \hfill
    \begin{subfigure}[b]{0.36\linewidth}
        \centering
        \includegraphics[width=\linewidth]{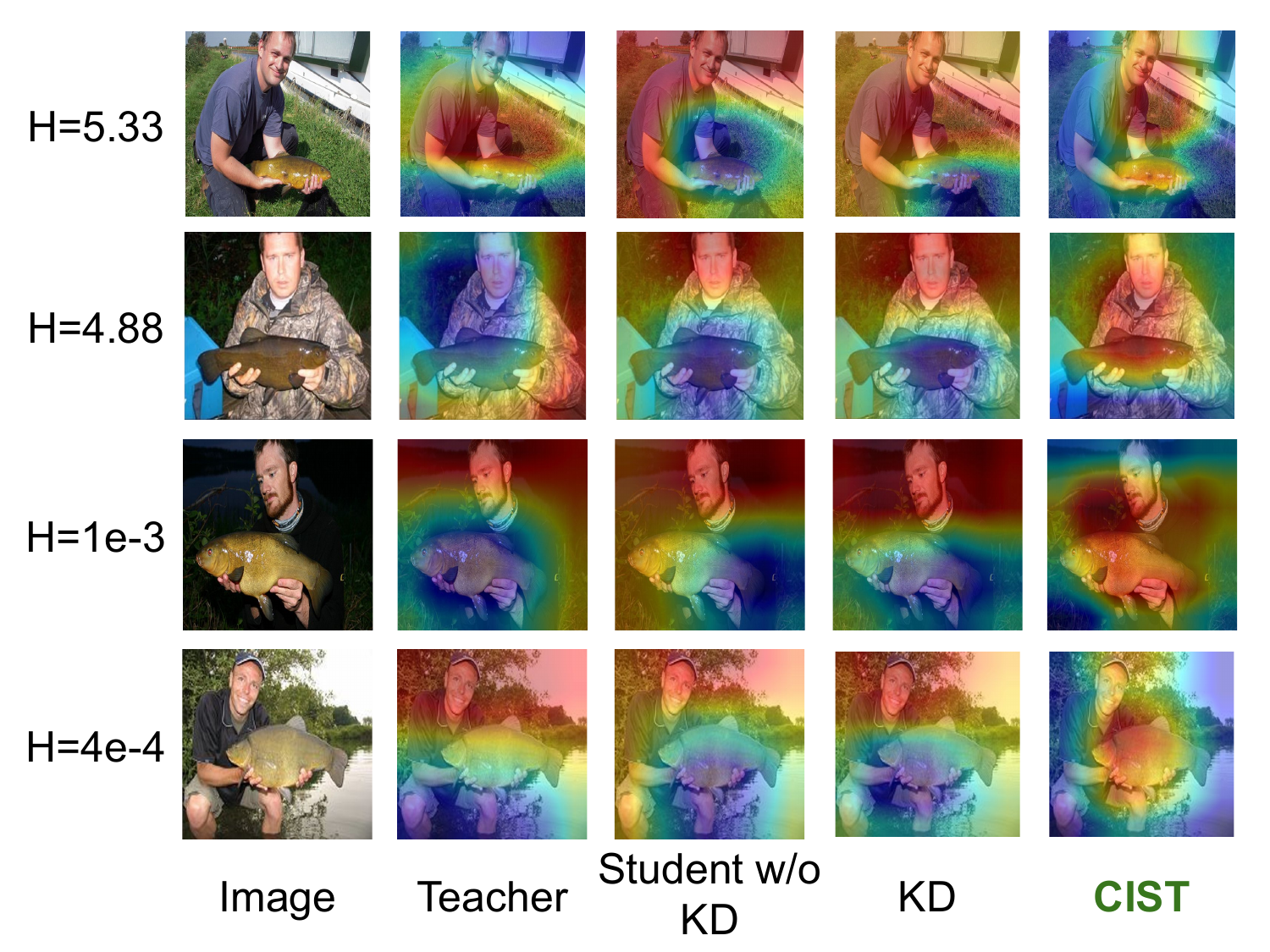}
        \vspace{-0.4cm}
        \caption{Grad-CAM}
        \label{fig:grad_cam}
    \end{subfigure}

    \vspace{-0.1cm}
    \caption{
    (a) t-SNE visualization of KD.
    (b) t-SNE visualization of CIST.
    (c) Grad-CAM visualizations for the teacher, student without KD, student with KD, and student with CIST on high- and low-entropy samples. The number on the left indicates the entropy value.
    }
    \vspace{-0.2cm}
    \label{fig:combined_tsne_gradcam}
\end{figure}

 \boldhdr{Training efficiency} 
We evaluate training efficiency by comparing the performance–cost trade-offs of different KD methods. As shown in Figure~\ref{fig:training_efficiency}, CIST achieves the best balance between accuracy and training time. While feature-based methods such as ReviewKD, CRD, and OFD incur additional overhead from auxiliary modules, CIST preserves the simplicity and efficiency of vanilla KD. All training times are measured on an NVIDIA RTX 3090 GPU.

\boldhdr{Distillation with ViT models and larger teachers}
We further evaluate CIST on Vision Transformer (ViT) backbones and larger teachers, increasing architectural diversity and teacher--student capacity gaps (see Appendix~\ref{appendix:more_exp}). CIST consistently improves vanilla KD by a clear margin, showing that its effectiveness extends beyond CNN-based models and remains strong with larger teachers.

\boldhdr{Grad-CAM} 
We select two high-entropy and two low-entropy samples based on the soft labels produced by a pretrained ResNet34 teacher, drawn from class 0 in the ImageNet dataset. Grad-CAM attribution maps are visualized for the teacher, a student trained without KD, a student trained with standard KD, and a student trained with CIST. Warm colors (e.g., red and yellow) indicate discriminative regions, while cool colors (e.g., blue and purple) highlight less informative areas. As shown in Figure~\ref{fig:grad_cam}, two key observations emerge: (1) the student trained with standard KD produces attribution maps that resemble those of the non-KD student rather than the teacher, suggesting that fixed-temperature KD fails to transfer spatially meaningful supervision; and (2) the student trained with CIST better focuses on the target object (e.g., the fish-tench), demonstrating that CIST provides more informative soft labels and enables more effective knowledge transfer.

\begin{wraptable}{r}{0.5\textwidth}
\vspace{-4mm}
\centering
\footnotesize
\caption{Importance of CIST components.}
\label{tab:ats_ablation}
\vspace{-2mm}
\begin{tabular}{l|c|c|c}
\toprule
Method & Reweight. & Temp. & Acc. \\
\midrule
KD &  &  & 73.33 \\
w/o Temp. & \checkmark &  & 74.41 \\
w/o Reweight. &  & \checkmark & \underline{76.49} \\
CIST & \checkmark & \checkmark & \textbf{76.96} \\
\bottomrule
\end{tabular}
\vspace{-4mm}
\end{wraptable}

\boldhdr{Effect of CIST components}
Table~\ref{tab:ats_ablation} reports a component-wise ablation of CIST. Confidence-aware reweighting alone improves standard KD to $74.41\%$, showing that sample-wise weighting provides useful curriculum regularization. CIST contributes a larger gain: removing the reweighting term while keeping adaptive temperatures reaches $76.49\%$, improving KD by $+3.16\%$. This confirms that stabilizing teacher-label entropy is the primary driver of CIST.
The full CIST objective achieves the best accuracy.

\vspace{-0.2cm}
\section{Conclusion}
\vspace{-0.2cm}

In this work, we analyze temperature scaling in knowledge distillation and show that the standard fixed-temperature design leads to inconsistent teacher-label entropy and rigid teacher--student logit-scale alignment. We propose CIST, a sample-wise adaptive temperature-scaling framework that assigns separate temperatures to the teacher and student and reweights distillation according to teacher confidence and student learning difficulty. Guided by our entropy analysis, CIST produces consistently informative soft labels, relaxes restrictive logit-scale matching, and emphasizes reliable samples. Experiments on both vision and language tasks show that CIST consistently outperforms standard KD and strong distillation baselines while adding negligible computational overhead.

\bibliography{main}
\bibliographystyle{plainnat}


\newpage

\appendix
\section{Implementation Details}\label{appendix:implementationdetails}

\boldhdr{Implementation} 
CIST is computationally efficient and easy to implement. It requires computing the maximum logit $v_{i, \max}$ per sample and dividing it by $\rho$ to obtain the temperature $\tau_i$. To preserve the shift-invariance property of softmax, each logit vector is centered by subtracting its mean: for a logit vector $\mathbf{x}$, we use $\mathbf{x} - \bar{\mathbf{x}}$, where $\bar{\mathbf{x}}$ is the mean of $\mathbf{x}$. Following the temperature-scaling principle of \citet{hinton2015distilling}, where $\tau>1$ smooths the output distribution and $\tau=1$ recovers the original prediction without additional smoothing, we define the adaptive temperature as
$
    \tau_i = \max(\frac{v_{i,\max}}{\rho}, 1).
$
For reproducibility, we provide the pseudo-code in Algorithm~\ref{algo:at}. Note that $\lambda_{\mathrm{CE}}$ and $\lambda_{\mathrm{KL}}$ in Algorithm~\ref{algo:at} are standard hyperparameters in traditional KD~\citep{hinton2015distilling}, rather than additional hyperparameters introduced by our method.

\boldhdr{How to set $\rho$}
The hyperparameter $\rho$ controls the target entropy of the softened predictions. To select $\rho$, we estimate the practical entropy range for a given dataset and teacher model using a small calibration subset, e.g., 512 training samples. The maximum entropy is $\log K$, where $K$ is the number of classes; for ImageNet-1k, $\log(1000)\approx 6.9$, corresponding to a uniform distribution. The lower end can be approximated by the entropy of the original teacher predictions before temperature scaling.
We then sweep candidate values of $\rho$ and choose an intermediate region, as shown in Figure~\ref{fig:entropy_rho}. Small $\rho$ values induce larger temperatures, pushing softened distributions toward high-entropy, nearly uniform predictions. Large $\rho$ values yield temperatures close to $1$, producing sharp predictions with limited dark knowledge. This process is performed once before training, so its computational overhead is negligible.

\boldhdr{CIST hyperparameter setting}
CIST replaces the fixed temperature hyperparameter with an entropy-control parameter $\rho$, which determines the target level of soft-label smoothing. We tune $\rho$ and the KD loss weight $\lambda_{\mathrm{KL}}$ by grid search, as shown in Figure~\ref{fig:alpha_analysis}. Based on this analysis, we set $\rho=3$ and $\lambda_{\mathrm{KL}}=8$ as the default configuration, which provides strong and stable performance across settings.

\boldhdr{Training resources}
All vision experiments were conducted on an Ubuntu Linux machine equipped with an NVIDIA RTX 3090 GPU with 24GB memory. Language experiments were conducted on a GPU cluster with four NVIDIA A100 GPUs, each with 40GB memory.

\subsection{Vision Task} 

We adopt standard training protocols widely used in prior distillation studies~\citep{zhao2022decoupled,jin2023multi,sun2024logit}. We use SGD with momentum and employ the original loss weights of each baseline method. Reported results are averaged over three independent runs. 

\boldhdr{Evaluation results}
We report Top-1 test accuracy (\%) for CIFAR-100 and Top-1 test accuracy (\%) and Top-5 test accuracy (\%) for ImageNet, computed as the classification accuracy on the test set.

\boldhdr{CIFAR-100}
All CIFAR-100 experiments are trained for $240$ epochs with batch size $64$, except MLKD~\citep{jin2023multi}, which follows its original $480$-epoch schedule. We use SGD with momentum 0.9 and weight decay $5\times 10^{-4}$. The initial learning rate is set to 0.01 for lightweight students (MobileNet~\citep{howard2017mobilenets,sandler2018mobilenetv2}, ShuffleNet~\citep{zhang2018shufflenet}) and 0.05 for other architectures (ResNet~\citep{he2016deep}, WRN~\citep{zagoruyko2016wide}, VGG~\citep{simonyan2014very}). The learning rate is decayed by a factor of $0.1$ at epochs $150$, $180$, and $210$. The cross-entropy (CE) loss weight is kept identical to the original baselines ($0.1$ for KD~\citep{hinton2015distilling} and CTKD~\citep{li2023curriculum}; $1.0$ for DKD~\citep{zhao2022decoupled} and MLKD~\citep{jin2023multi}). 

\boldhdr{ImageNet}
For ImageNet, we train for 100 epochs using SGD with batch size $512$, momentum $0.9$, and weight decay $10^{-4}$. The initial learning rate is $0.2$ and is divided by $10$ at epochs $30$, $60$, and $90$. We keep the baseline CE loss weight unchanged ($0.1$ for KD and CTKD; $0.5$ for the remaining baselines). For CIST, we do not have the temperature hyperparameter and instead use entropy control $\rho=3$ with KD loss weight $\lambda=8$ by default.

\boldhdr{Setting for Table~1}
For fair comparison, KD-EntOut-CE and KD-EntOut-HT use the same hyperparameters as standard KD such as CE loss weight $0.1$ and KL loss weight $0.9$.

\boldhdr{Combining CIST with other distillation objectives}
We evaluate CIST on top of MLKD~\citep{jin2023multi}. MLKD performs logit alignment at multiple granularities (instance, batch and class level) via temperature-scaled softmax at each level. We therefore integrate CIST by replacing the original temperature scaling in MLKD with our entropy-guided, sample-wise temperature scaling CIST.

We do not apply CIST to CTKD~\citep{li2023curriculum} as CTKD optimizes a learnable global temperature parameter, while CIST eliminates shared temperature scaling and assigns entropy-guided sample-wise temperatures. Hence, the two formulations are not directly compatible without modifying CTKD.

\subsection{Language Task}

\boldhdr{Evaluation results}
We focus on the off-policy distillation setting, where training is performed using fixed teacher-generated responses. Therefore, we do not compare with methods that require on-policy sampling or additional external datasets. All methods are trained under the same setup, using the same dataset, identical hyperparameters, a shared teacher checkpoint, and the same student initialization to ensure a fair comparison. 
For evaluation, checkpoints are saved after each epoch, and we report results from the checkpoint with the best validation ROUGE-L. ROUGE-L is averaged over five random seeds $\{10,20,30,40,50\}$, and the decoding temperature is set to $1$ by default.

\boldhdr{Hyperparameter setting}
Following~\citet{gu2023minillm}, we use AdamW with a weight decay of $0.01$. The learning rate is set to $5\times10^{-4}$ for the GPT-2 0.1B student and $5\times10^{-5}$ for OPT-1.3B. We use a batch size of $32$ and train each student model for $20$ epochs. The maximum input length is fixed to $512$ tokens for all model families.

Motivated by prior studies showing that RKL often outperforms FKL in language model distillation~\citep{chan2022greedification,gu2023minillm, kim2024promptkd, wu2025rethinking, luong2026diversity}, we apply CIST to RKL in our language experiments.
For baseline-specific hyperparameters, we follow the hyperparameter settings from the original works and use the implementation provided in DistilLM~\citep{ko2024distillm}: Sym-KL which is $0.5\,\mathrm{FKL} + 0.5\,\mathrm{RKL}$, SRKL use smoothing coefficient $\alpha=0.1$~\citep{ko2024distillm}, while $\alpha$-$\beta$ divergence uses $(\alpha,\beta)=(0.2,0.7)$~\citep{wang2025abkd}.

\begin{algorithm}[t]
\caption{Consistently Informative Soft-label Temperature (CIST)}
\begin{algorithmic}[1]
\Require Train set $\mathcal{D} = \{( \mathbf{x}_n, y_n)\}_{n=1}^N$, teacher model $f_T$, student model $f_S$, number of epochs $E$, entropy control constant $\rho$, CE loss weight $\lambda_{\rm CE}$, KL loss weight $\lambda_{\rm KL}$.
\Ensure Trained student model
\For{$\text{epoch} = 1,2,\dots,E$} 
    \ForAll{$(\mathbf{x}_n, y_n)$ in $\mathcal{D}$}
        \State $\mathbf{v}_n = f_T(\mathbf{x}_n), \ \ \mathbf{z}_n = f_S(\mathbf{x}_n)$
        \State $\hat{ \mathbf{v} }_n = \mathbf{v}_n - \bar{\mathbf{v}}_n, \ \ \hat{ \mathbf{z} }_n = \mathbf{z}_n - \bar{\mathbf{z}}_n$
        \State $v_{\max} = \max(\hat{\mathbf{v}}_n), \ \ z_{\max} = \max(\hat{\mathbf{z}}_n),$
        \State $\tau^t_n \leftarrow \max ( v_{\max} / \rho, 1), \ \ \tau^s_n \leftarrow \max ( z_{\max} / \rho, 1)$
        \State $\mathbf{p}_n = \mathrm{softmax} (\hat{\mathbf{v}}_n / \tau^t_n), \ \ \mathbf{q}_n = \mathrm{softmax}(\hat{\mathbf{z}}_n / \tau^s_n)$
        \State Update student by minimizing $\lambda_{\text{CE}} \mathcal{L}_{\text{CE}} ( y_n, \mathbf{q}_n) + \lambda_{\text{KL}}\tau^t_n \tau^s_n \mathcal{L}_{\rm KL}(\mathbf{p}_n \| \mathbf{q}_n)$
    \EndFor
\EndFor
\end{algorithmic}
\label{algo:at}
\end{algorithm}

\section{Proof of Proposition~\ref{prop:ats}}\label{appendix:proof}

\boldhdr{Proof}  
Substituting $\mathbf{p}^{\tau_i}_i$ as the softmax output for a given sample $i$ with temperature $\tau_i$ into entropy Eq.~(\ref{eq:entropy}) and using the property that $\sum^{C}_{c=1} p_{i, c}^{\tau_i} = 1$, we can rewrite it as follows
\begin{align}
    H\left( \mathbf{ p }^{\tau_i}_i \right) 
    &= \log \sum^{C}_{c=1} \exp\left( \frac{ v_{i, c} } { \tau_i }\right) - \sum_{c=1}^{C} p^{\tau_i}_{i,c} \frac{ v_{i, c} }{ \tau_i }.
\end{align}
Considering two samples $i$ and $j$, the magnitude of entropy difference between their soft labels can be expressed as:
\begin{align}
    &\left| H( \mathbf{ p }^{\tau_i}_i ) - H ( \mathbf{ p }^{\tau_j}_j ) \right| \nonumber \\
    &=\left| \log \frac{ \small{ \sum^{C}_{c=1} } \exp( \frac{ v_{i, c} } { \tau_i } ) } { \small{ \sum^{C}_{c=1} } \exp( \frac{ v_{j, c} } { \tau_j } ) } + \small{ \sum_{c=1}^{C} }\big( p^{\tau_i}_{j,c} \frac{ v_{j, c} }{ \tau_j } - p^{\tau_j}_{i,c} \frac{ v_{i, c} }{ \tau_i } \big) \right|.
    \label{eq:entropy_diff}
\end{align}
Let $m = \arg \max_c z_{i, c}$ and $m' = \arg \max_c z_{j, c}$ be the indices of the maximum logits for samples $i$ and $j$, respectively. The first term, denoted as $A$, simplifies to:
\begin{align*}
    A 
    &= \log \frac{ \exp( \frac{ v_{i,m} }{ \tau_i } ) \left( 1 + \sum^{C}_{c\not=m} \exp( \frac{ v_{i,c} - v_{i, m} }{ \tau_i }  )\right) } { \exp( \frac{ v_{j,m'} }{ \tau_j } ) \left( 1 + \sum^{C}_{c\not=m'} \exp( \frac{ v_{j,c} - v_{j, m'} }{ \tau_j }) \right) } \\
    &= \log \frac{ \exp( \frac{ v_{i,m} }{ \tau_i } ) } { \exp( \frac{ v_{j,m'} }{ \tau_j } )} + \log \frac{1 + \sum^{C}_{c\not=m} \exp( \frac{ v_{i,c} - v_{i, m} }{ \tau_i } ) }{1 + \sum^{C}_{c\not=m'} \exp( \frac{ v_{j,c} - v_{j, m'} }{ \tau_j })}.
\end{align*}
Since the teacher is well-trained, the maximum logit dominates for both samples $i$ and $j$, i.e., $v_{i, m} \gg v_{i, c}, \forall c\not=m$, or equivalently, $v_{i,c} - v_{i,m}$ becomes very negative. The same applies to sample $j$. Therefore, the second term above is negligible, as the exponential of a large negative value approaches zero. Then, term $A$ is approximated by the difference between the two dominant logits:
\begin{align}
    A \approx \log \frac{ \exp( \frac{ v_{i,m} }{ \tau_i } ) } { \exp( \frac{ v_{j,m'} }{ \tau_j } )} = \frac{v_{i,m}}{\tau_i} - \frac{v_{j, m'}}{\tau_j}.
\end{align}

We now analyze the second term in Eq.~(\ref{eq:entropy_diff}), denoted as $B$. Let $\pi_i, \pi_j$ be permutations that sort the logits of samples $i$ and $j$ in decreasing order, such that index $1$ corresponds to the maximum logit and index $C$ to the minimum. Then:
\begin{align}
    B 
    &= \sum_{c=1}^{C} \left( p^{\tau}_{j, \pi_j(c)} \frac{ v_{j, \pi_j(c)} }{ \tau_j } - p^{\tau}_{i, \pi_i(c)} \frac{ v_{i, \pi_i(c)} }{ \tau_i } \right).
\end{align}
The value of $B$ is primarily influenced by the difference between the maximum logits, $v_{i, \pi_i(1)} = v_{i, m}$ and $v_{j, \pi_j(1)} = v_{j, m'}$, as well as their corresponding softened probabilities, $p^{\tau}_{i, m}$ and $p^{\tau}_{j, m'}$, which tend to be highly peaked in well-trained models. To reduce both terms $A$ and $B$, we propose to adaptively adjust the temperatures $\tau_i$ and $\tau_j$ so that the ratio between each sample's maximum logit and its temperature is normalized to a predefined constant $\rho$:
\begin{align}
\label{eq:motivation}
    \frac{ v_{i,m} } { \tau_i } = \frac{ v_{i,m'} } { \tau_j} = \rho.
\end{align}
Under this scaling, the term $A \approx 0$ as the dominant logits are aligned. The value of $B$ is also reduced, since its leading difference components become aligned across the two samples. The remaining variation arises from differences in the smaller logits, which are not significant. As a result, the overall entropy gap $| H( \mathbf{ p }^{\tau_i}_i ) - H( \mathbf{ p }^{\tau_j}_j ) |$ is minimized, which completes the proof. 





\section{More Experiments}
\label{appendix:more_exp}

\begin{table}[t]
\centering
\small
\setlength{\tabcolsep}{3pt}
\renewcommand{\arraystretch}{1.15}
\caption{Top-1 test accuracy (\%) of student WRN-16-2 distilled from various teachers on CIFAR-100. Best results are in \textbf{bold}.}
\label{tab:various_tea}
\scalebox{0.9}{
\begin{tabular}{l|cccccc}
\toprule
\multirow{2}{*}{Method} 
& \multicolumn{6}{c}{Teacher Model} \\
\cmidrule(lr){2-7}
& VGG13 & WRN-28-2 & WRN-40-2 & WRN-16-4 & WRN-28-4 & ResNet50 \\
\midrule
Teacher Acc. 
& 74.64 & 75.45 & 75.61 & 77.51 & 78.60 & 79.34 \\
\midrule
KD     & 74.93 & 75.37 & 74.92 & 75.79 & 75.04 & 75.36 \\
DKD    & 75.45 & 75.92 & 76.24 & 76.00 & 76.45 & \textbf{76.60} \\
KD+LS  & 75.03 & \textbf{76.32} & 76.11 & 76.72 & 75.77 & 76.24 \\
\midrule
CIST (Ours) 
& \textbf{75.52} & 76.08 & \textbf{76.59} & \textbf{76.87} & \textbf{76.48} & \textbf{76.60} \\
\bottomrule
\end{tabular}
}
\end{table}

\boldhdr{Distillation with large teacher} 
Prior works~\citep{cho2019efficacy,huang2022knowledge} show that larger teachers do not always lead to better distillation performance due to capacity gaps with lightweight students. As discussed in our analysis, CIST addresses this issue by relaxing strict logit matching and promoting relative scale alignment. As shown in Table~\ref{tab:various_tea}, CIST significantly improves student performance over traditional KD and consistently achieves the best or second-best accuracy across all six teacher models.

\begin{table}[t]
\footnotesize
\renewcommand\arraystretch{0.9} 
\centering
\caption{Top-1 accuracy (\%) of KD methods on CIFAR-100. The teacher is ResNet56.}
\label{tab:vit}
\scalebox{1.05}{
\begin{tabular}{l|c|ccc}
\toprule
Student & Params & Train & KD & KD+Ours \\
\midrule
DeiT-Ti   & 5M & 65.08 & 73.25 & \textbf{77.04} \\
T2T-ViT-7 & 4M & 69.37 & 74.15 & \textbf{74.53} \\
PiT-Ti    & 5M & 73.58 & 75.47 & \textbf{77.83} \\
\bottomrule
\end{tabular}
}
\end{table}

\boldhdr{ViT experiments}
To assess the generality of CIST beyond CNN backbones, we evaluate it on Transformer-based architectures via knowledge distillation on four representative Vision Transformer (ViT) variants: DeiT-Ti~\citep{pmlr-v139-touvron21a}, T2T-ViT-7~\citep{Yuan_2021_ICCV}, and PiT-Ti~\citep{Heo_2021_ICCV}. We follow the experimental protocol of prior work~\citep{sun2024logit} and conduct all experiments on CIFAR-100, using the same training pipeline and hyperparameter settings to ensure fair comparison. As reported in Table~\ref{tab:vit}, CIST (KD+Ours) consistently improves over vanilla KD across all ViT backbones, indicating that CIST is effective on Transformer architectures and not limited to CNN-based distillation.

\section{Limitations and Discussions}
\label{appendix:limitations}

CIST is designed for logit-based distillation, where KD matches softened output distributions. Thus, it naturally applies to classification tasks and can be extended to classification heads in detection or segmentation. However, applying CIST to regression heads or other non-logit outputs is non-trivial, since temperature scaling and soft-label entropy are not directly defined in these settings.

CIST also depends on teacher-logit quality. When the teacher is weak, poorly calibrated, or unreliable on many samples, its logits may provide poor confidence signals, limiting the benefit of adaptive temperature scaling. This is a potential failure case, although it falls outside the standard KD assumption that the teacher is stronger than the student.

Due to computational constraints, our language experiments use teacher models up to 7B parameters, and our vision experiments focus on CIFAR-100 and ImageNet. Evaluating CIST on larger models and broader vision-language tasks remains future work.
Although on-policy sampling, stronger student initialization, auxiliary supervision, and task-specific enhancements may further improve performance, we exclude them to ensure a direct and fair comparison between distillation objectives under the same controlled setting.


\newpage

\end{document}